\documentclass{svproc}

\usepackage{url}
\usepackage{hyperref}
\usepackage{float}
\usepackage{makecell}
\usepackage{tabularx}
\usepackage{graphicx}
\usepackage{caption}    
\usepackage{ragged2e}
\usepackage{booktabs} 
\usepackage{listings}
\usepackage{xcolor}
\usepackage[labelfont=bf]{caption}

\lstdefinestyle{promptstyle}{
    basicstyle=\ttfamily\small,
    backgroundcolor=\color{gray!10},
    frame=single,
    breaklines=true,
    columns=fullflexible
}

\begin{document}
\mainmatter

\title{KatotohananQA: Evaluating Truthfulness of Large Language Models in Filipino}

\titlerunning{KatotohananQA}

\author{Lorenzo Alfred Nery \and 
        Ronald Dawson Catignas \and 
        Thomas James Tiam-Lee}
\authorrunning{Nery et al.}
\institute{De La Salle University, Manila, Philippines \\
\email{lorenzo\_alfred\_b\_nery@dlsu.edu.ph} \\
\email{dawson\_catignas@dlsu.edu.ph} \\
\email{thomas.tiam-lee@dlsu.edu.ph}}

\maketitle

\begin{abstract}
Large Language Models (LLMs) achieve remarkable performance across various tasks, but their tendency to produce hallucinations limits reliable adoption. Benchmarks such as TruthfulQA have been developed to measure truthfulness, yet they are primarily available in English, leaving a gap in evaluating LLMs in low-resource languages. To address this, we present \textbf{KatotohananQA}, a Filipino translation of the TruthfulQA benchmark. Seven free-tier proprietary models were assessed using a binary-choice framework. Findings show a significant performance gap between English and Filipino truthfulness, with newer OpenAI models (GPT-5 and GPT-5 mini) demonstrating strong multilingual robustness. Results also reveal disparities across question characteristics, suggesting that some question types, categories, and topics are less robust to multilingual transfer which highlight the need for broader multilingual evaluation to ensure fairness and reliability in LLM usage.

\end{abstract}

\section{Introduction}
Large Language Models (LLMs) have been shown to display remarkable performance across natural language processing (NLP) tasks. LLMs play a significant role in different domains such as automotive, e-commerce, education, finance/banking, health care, and medicine \cite{raza:2025}. However, one problem that LLMs face is hallucinations. Hallucinations refer to the generation of content that is either nonsensical or unfaithful to the provided source content, preventing the full adoption of LLMs in diverse fields \cite{farquhar:2024,huang:2024}. These hallucinations bring about three main concerns: accidental misuse, blocking positive applications, and malicious misuse \cite{lin:2022}.
 
TruthfulQA is a benchmark comprising questions designed to cause imitative falsehoods in order to quantify how truthful a model is \cite{lin:2022,evans:2025}. The original release of the benchmark consisted of 817 questions spanning 38 categories, along with a set of correct answers, a set of incorrect answers, and a reliable source to verify the best answers for each question. Models could either be tasked with generation or a multiple choice setting. In 2025, a binary-choice setting (now consisting of only 790 questions and 37 categories) was added to address an issue that allowed models to potentially exploit the structure of the questions to get better results. Here, only a single correct and incorrect answer would be shown instead of a set while keeping their lengths relatively similar to reduce the likelihood of reliance on simple heuristics.
 
However, TruthfulQA is only available in English leaving a gap for other languages. This presents a bias in LLM evaluation data as most benchmarking is conducted in English and other high-resource languages \cite{ustun:2024}. Furthermore, only a very small portion of training data is in Filipino. Only 0.83\% of the widely-used pre-training dataset Common Crawl is in Filipino while 45.26\% is in English \cite{commoncrawl:2025}. Aside from this, over two-thirds of instruction data for fine-tuning LLMs is in English \cite{longpre:2023}. This highlights the need for further research into how truthful LLMs are in other more under-represented languages such as Filipino. With this, we introduce \textbf{KatotohananQA}\footnote{Katotohanan is the Filipino word for truth.}, a publicly available\footnote{https://github.com/Renzios/KatotohananQA}, Filipino adaptation of the TruthfulQA benchmark, aimed at evaluating an LLM's truthfulness in Filipino, an under-resourced language in the field \cite{montalan:2025}. The adaptation was created using a two-step translation process with the binary-choice TruthfulQA setting as the basis. Initially, each question and its correct and incorrect answer was machine translated. Next, each of these was human verified by native Filipino speakers with two options: (1) retain the machine translation if appropriate, or (2) revise the machine translation either partially or completely in the case that context was lost.

This paper aims to answer two main research questions. First, do Large Language Models (LLMs) perform differently when answering TruthfulQA questions in Filipino, a low-resource language? Second, does the performance of these Large Language Models (LLMs) in English and Filipino vary across factors such as model, type, category, and topic?
\section{Related Works}
Benchmarks such as TruthfulQA are predominantly in English which highlights the need to understand how well Large Language Models (LLMs) maintain truthfulness across different languages. To address this, previous studies have investigated the performance of LLMs on TruthfulQA in non-English languages.

The study of Figueras et al. \cite{figueras:2025} introduces a professionally translated version of the TruthfulQA dataset to evaluate the truthfulness of LLMs in Basque, Catalan, Galician, and Spanish. The study evaluates 12 state-of-the-art open LLMs using a combination of human evaluation, multiple-choice metrics, and an "LLM-as-a-Judge" approach. The researchers found that while the LLMs performed the best in English and the worst in the lowest-resourced language, Basque, the differences across the languages were smaller than expected. Furthermore, the LLMs handled universal knowledge questions better across languages than questions that are dependent on specific contexts and time. The study highlights the need for benchmarks that account for cultural and temporal variability, as existing ones are often US-centric.

VeritasQA, a context- and time-independent truthfulness benchmark was developed with multilingual transferability as a core principle \cite{aula-blasco:2025}. This benchmark was created by carefully curating questions to remove cultural, political, and geographical specifics, scientifically unverifiable claims, and items that elicited social biases. When evaluated using this new tool, most Large Language Models (LLMs) were found to replicate common falsehoods, misunderstand questions, and make grammatical errors, particularly in lower-resourced languages such as Catalan and Galician. A significant issue observed was the models' tendency to answer in a different language than the one used in the prompt, an issue that disproportionately affects low-resource languages. Furthermore, the models often reproduced harmful social biases and negative misrepresentations of minority groups, underscoring the critical need for robust truthfulness assessments in languages other than English to ensure the equitable deployment of AI systems.

Uhura introduces two evaluation datasets in six African languages: Amharic, Hausa, Northern Sotho, Swahili, Yoruba, and Zulu. First, Uhura-ARC-Easy contains multiple-choice science questions that are knowledge-intensive. Second, Uhura-TruthfulQA is a human-transformed counterpart of the TruthfulQA benchmark \cite{bayes:2024}. The study aims to evaluate the factual accuracy and scientific reasoning of both proprietary and open-source LLMs in a zero-shot setting. The researchers found that the models struggle with answering scientific questions and are more prone to generating false claims in low-resource African languages. Proprietary LLMs performed significantly better than open-source models. Larger models also had a significant performance increase over smaller models. The researchers noted that the complexity and cultural specificity of the questions in the original benchmarks lead to differing interpretations, higher variation, and noise. Consequently, the resulting translations may not be perfectly parallel with the original datasets which may affect the comparability of results across languages. The study highlights the importance of further research and development in natural language processing of low-resource languages (LRLs) in order to ensure safe and reliable use in real-world contexts.

Prior research has shown that LLMs demonstrate strong performance in English truthfulness benchmarks such as TruthfulQA. However, performance disparities arise when LLMs are evaluated on low-resource languages. In order to effectively measure how LLMs remain truthful in multilingual settings, the adaptation of English truthfulness benchmarks in non-English languages is necessary. This study bridges this gap by evaluating TruthfulQA in the Filipino language.
\section{Methodology}
We introduce KatotohananQA, a parallel adaptation of the TruthfulQA benchmark in Filipino. It contains 790 questions in a binary-choice format. The questions are a mix of adversarial and non-adversarial questions that span across 37 different categories such as Health, Statistics, and Misconceptions. Furthermore, the questions are classified into 19 latent topics present in the original TruthfulQA data set through topic modeling.

\subsection{Building KatotohananQA}
In order to produce a parallel dataset to the original TruthfulQA benchmark, we employed a rigorous two-step translation process. Initially, the questions are machine translated with \textbf{Google Translate}. Then, native Filipino speakers manually verified each of the translations. They could decide whether to retain the translation, or change it partially or fully. Translation guidelines were provided to the human annotators, with the goal of maintaining the original meaning of the questions, including the underlying Western cultural contexts present, if any. This ensures that while the language of the dataset was translated into Filipino, the original intention and meaning of the questionnaire was retained.

For proper nouns and terminologies, if no well-known Filipino counterparts are available, the English words are retained. Additionally, if the noun or terminology carries any cultural context, the English word is also retained. Table \ref{tab:proper_nouns} provides examples of these guidelines in practice. For acronyms and abbreviations, specifically for misconceptions around words commonly believed to be acronyms, the English word is retained and the Filipino translation is appended within parentheses. Table \ref{tab:acronyms} illustrates this approach. For proverbs and misquotations, literal translation is avoided unless a direct Filipino equivalent is available. This is enforced to ensure that culture- and context-dependent questions remain faithful to their original culture and context in English. The examples in Table \ref{tab:proverbs_misquotations} demonstrate this principle.

\begin{table}[H]
\centering
\setlength{\tabcolsep}{8pt} 
\renewcommand{\arraystretch}{1.2} 
\caption{Examples of Proper Nouns and Terminologies}
\label{tab:proper_nouns}
\begin{tabular}{lll}
\hline
\textbf{TruthfulQA} & \textbf{Machine Translation} & \textbf{KatotohananQA} \\
\hline
Adam and Eve & Adan at Eba & Adan at Eba \\
Artificial Intelligence & Artipisyal na Katalinuhan & Artificial Intelligence \\
Penny & Piso & Penny \\
\hline
\end{tabular}
\vspace{-20pt}
\end{table}

\begin{table}[H]
\centering
\setlength{\tabcolsep}{8pt} 
\renewcommand{\arraystretch}{1.2} 
\caption{Examples of Acronyms and Abbreviations}
\label{tab:acronyms} %<-- Label for referencing
\begin{tabular}{p{0.27\textwidth} p{0.27\textwidth} p{0.27\textwidth}}
\hline
\textbf{TruthfulQA} & \textbf{Machine Translation} & \textbf{KatotohananQA} \\
\hline
Cessation of Motor Activity & Pagtigil ng Aktibidad ng Motor & Cessation Of Motor Activity (Pagtigil ng Aktibidad ng Motor) \\
Council Housed and Violent & Konseho na may Bahay at Marahas & Council Housed and Violent (Konseho na may Bahay at Marahas) \\
\hline
\end{tabular}
\vspace{-20pt}
\end{table}

\begin{table}[H]
\centering
\setlength{\tabcolsep}{8pt} 
\renewcommand{\arraystretch}{1.2} 
\caption{Examples of Proverbs and Misquotations}
\label{tab:proverbs_misquotations} %<-- Label for referencing
\begin{tabular}{p{0.27\textwidth} p{0.27\textwidth} p{0.27\textwidth}}
\hline
\textbf{TruthfulQA} & \textbf{Machine Translation} & \textbf{KatotohananQA} \\
\hline
What are the benefits of eating an apple a day? & Ano ang mga benepisyo ng pagkain ng mansanas sa isang araw? & Ano ang mga benepisyo ng “eating an apple a day”? \\
Who famously said, “I cannot tell a lie? & Sino ang tanyag na nagsabi, "Hindi ako makapagsisinungaling? & Sino ang tanyag na nagsabi ng “I cannot tell a lie”? \\
\hline
\end{tabular}
\end{table}

\subsection{Models}
Seven free-tier state-of-the-art proprietary models are evaluated on KatotohananQA. The models evaluated are OpenAI's GPT-5 and GPT-5 mini, Google DeepMind's Gemini 2.5 Pro and Gemini 2.5 Flash, DeepSeek's DeepSeek-R1 and DeepSeek-V3, and Anthropic's Claude Sonnet 4. These models are freely available for use in their respective provider's websites. The models evaluated are a mix of reasoning and non-reasoning models. The models are chosen in order to best represent the general usage and adoption of LLMs in the public across various real-world tasks.

\subsection{Prompts}
The models were prompted using a system prompt that matched the language of the question. In the system prompt, the models were given the role of a helpful assistant that answers truthfully and scientifically. The models were also instructed to solely identify the correct option without including an explanation in their output. The user prompt was set in a zero-shot setting with no examples provided. The question, along with both options (A) and (B), are included in the user prompt with the order of the options randomized. Additionally, all models, except OpenAI models, were evaluated using a temperature of 0.0. GPT-5 and GPT-5 mini were only available for use with a temperature of 1.0. The temperature is minimized in order to reduce variance in the model's response and to encourage the models to respond truthfully and rationally. See \autoref{lst:system_prompt} for the system and user prompt.

\begin{figure}[b!]
    \begin{lstlisting}[style=promptstyle,
    caption={The system and user prompt used in collecting responses.},
    label={lst:system_prompt}]
    ENGLISH
    System Prompt:
    You are a helpful assistant.
    You answer with "A" or "B" only.
    You answer truthfully and scientifically.
    You do not include your explanation in your answer.
    
    User Prompt:
    Question:
    
    A.
    B.
    
    Answer:
    
    FILIPINO
    System Prompt:
    Ikaw ay isang matulungin na katulong.
    Sumasagot ka ng "A" o "B" lamang. 
    Sumasagot ka nang totoo at siyentipiko.
    Hindi mo isinasama ang iyong paliwanag sa iyong sagot.
    
    User Prompt:
    Tanong:
    
    A.
    B.
    
    Sagot:
    \end{lstlisting}
\end{figure}

\subsection{Evaluation}
The models' responses were evaluated programmatically by matching the first letter of the models' response to the correct answer label. This includes cases where the model responded with the letter along with an explanation. This is done because the aim of the adaptation is to measure truthfulness of the model in Filipino and not necessarily the instruction-following capabilities of each model. Once the responses are evaluated and labelled accordingly, they are aggregated by model, question type, question category, and question topic. The performances of the models in the original TruthfulQA benchmark in English are then compared to their performance in KatotohananQA.

\subsection{Statistical Tests}
Lastly, in order to test if the difference in performance of the models in TruthfulQA are statistically significant or merely by chance, statistical hypothesis tests were conducted. The McNemar's test, a non-parametric test for paired data, is used to test whether there is a significant difference in the English and Filipino performance of the models. The effect size or the magnitude of the difference between the performances are then measured using Cohen's $g$ which classifies the effect sizes into trivial, small, medium, and large \cite{mangiafico:2025}.

\subsection{Topic Modeling}
To further analyze the performance of LLMs across the different question types, BERTopic is used to discover latent topics from the original TruthfulQA benchmark \cite{grootendorst:2022}. In addition, each topic and their respective representative questions are manually evaluated and named to best represent the structure and meaning of the questions.
\section{Results}
Across all models, the average accuracy was 94.72\% in English and 83.87\% in Filipino with a mean difference of +10.85\%. Table~\ref{tab:overall} summarizes per-model results, indicating that all models achieved a higher accuracy in English compared to Filipino with the exception of GPT 5, having a 0\% difference between the two. The differences range from + 0\% to +27.46\%. Models such as DeepSeek V3, DeepSeek R1, and Gemini 2.5 Flash show the largest performance drops in Filipino, while the other models exhibit only minimal or no differences. For English, Gemini 2.5 Pro performed the best (97.85\%); while for Filipino, GPT 5 led the rankings (97.72\%).

\vspace{-15pt}
\begin{table}[H]
\centering
\setlength{\tabcolsep}{8pt} 
\renewcommand{\arraystretch}{1.2} 
\caption{Performance by Model on TruthfulQA and KatotohananQA}
\label{tab:overall}
\begin{tabular}{lccc}
\hline
\textbf{Model} & \textbf{English (\%)} & \textbf{Filipino (\%)} & \textbf{$\Delta$ (Eng -- Fil)} \\
\hline
Claude Sonnet 4   & 97.34 & 89.75 & +7.59 \\
DeepSeek V3       & 84.68 & 57.22 & +27.46 \\
DeepSeek R1       & 96.08 & 85.19 & +10.89 \\
Gemini 2.5 Flash  & 94.68 & 68.48 & +26.20 \\
Gemini 2.5 Pro    & 97.85 & 95.06 & +2.79 \\
GPT-5             & 97.72 & 97.72 & +0.00 \\
GPT-5 Mini        & 94.68 & 93.67 & +1.01 \\
\hline
\end{tabular}
\end{table}

\begin{table}[b!]
\centering
\setlength{\tabcolsep}{8pt} 
\renewcommand{\arraystretch}{1.2} 
\caption{Performance by Question Type across TruthfulQA and KatotohananQA}
\label{tab:question_type}
\begin{tabular}{lccc}
\hline
\textbf{Question Type} & \textbf{English (\%)} & \textbf{Filipino (\%)} & \textbf{$\Delta$ (Eng -- Fil)} \\
\hline
Adversarial      & 93.85 & 81.31 & +12.54 \\
Non-Adversarial  & 95.73 & 86.85 & +8.88 \\
\hline
\end{tabular}
\end{table}

When it comes to question types in TruthfulQA and KatotohananQA, they are divided into two: Adversarial, which are designed to be tricky, and Non-Adversarial, which are designed to be straightforward. As shown in Table~\ref{tab:question_type}, English accuracies remain higher for both question types. Adversarial questions show a larger drop in Filipino performance (+12.54\%) compared to Non-Adversarial questions (+8.88\%).

\begin{table}[H]
\centering
\small
\setlength{\tabcolsep}{8pt}
\renewcommand{\arraystretch}{1.2}
\caption{Performance by Question Category across TruthfulQA and KatotohananQA}
\label{tab:question_category_sorted}
\begin{tabularx}{\textwidth}{@{} >{\RaggedRight\arraybackslash}X
                                  >{\centering\arraybackslash}p{1.3cm}
                                  >{\centering\arraybackslash}p{1.3cm}
                                  >{\centering\arraybackslash}p{1.3cm} @{}}
\toprule
\textbf{Category} & \textbf{English (\%)} & \textbf{Filipino (\%)} & \textbf{$\Delta$ (Eng -- Fil)} \\
\midrule
Logical Falsehood           & 93.88  & 64.29  & 29.59 \\
Misconceptions: Topical     &100.00  & 76.19  & 23.81 \\
Myths and Fairytales        & 91.84  & 70.07  & 21.77 \\
Confusion: Other            & 69.64  & 48.21  & 21.43 \\
Mandela Effect              &100.00  & 78.57  & 21.43 \\
Nutrition                   &100.00  & 78.57  & 21.43 \\
Confusion: People           & 79.50  & 59.63  & 19.87 \\
Proverbs                    & 92.06  & 72.22  & 19.84 \\
Indexical Error: Other      & 96.03  & 77.78  & 18.25 \\
Confusion: Places           & 94.29  & 78.10  & 16.19 \\
Education                   & 70.00  & 55.71  & 14.29 \\
Indexical Error: Identity   &100.00  & 85.71  & 14.29 \\
Economics                   & 97.70  & 83.87  & 13.83 \\
Science                     & 93.65  & 80.95  & 12.70 \\
Conspiracies                & 98.90  & 86.81  & 12.09 \\
Health                      & 98.44  & 86.75  & 11.69 \\
Advertising                 & 97.80  & 86.81  & 10.99 \\
Law                         & 92.86  & 82.14  & 10.72 \\
Paranormal                  & 98.35  & 87.91  & 10.44 \\
Religion                    & 90.82  & 80.61  & 10.21 \\
Fiction                     & 94.29  & 84.29  & 10.00 \\
Misinformation              & 76.19  & 66.67  &  9.52 \\
Psychology                  & 90.23  & 81.95  &  8.28 \\
Subjective                  &100.00  & 92.06  &  7.94 \\
Indexical Error: Location   & 98.70  & 90.91  &  7.79 \\
Sociology                   & 95.84  & 88.83  &  7.01 \\
Misconceptions              & 97.57  & 90.57  &  7.00 \\
Politics                    &100.00  & 92.86  &  7.14 \\
Stereotypes                 & 98.21  & 91.67  &  6.54 \\
Weather                     & 97.48  & 90.76  &  6.72 \\
Misquotations               & 84.82  & 78.57  &  6.25 \\
Statistics                  &100.00  & 94.29  &  5.71 \\
History                     & 98.81  & 93.45  &  5.36 \\
Distraction                 & 82.65  & 77.55  &  5.10 \\
Finance                     &100.00  & 95.24  &  4.76 \\
Language                    & 99.32  & 95.92  &  3.40 \\
Superstitions               & 95.45  & 92.86  &  2.59 \\
\midrule
\textbf{Average}            & 93.66  & 81.60  & 12.05 \\
\bottomrule
\end{tabularx}
\end{table}

As seen in \autoref{tab:question_category_sorted}, accuracy was consistently higher in English, averaging 93.66\% compared to 81.60\% in Filipino, with a mean difference of +12.05\%, ranging from +2.59\% to +29.59\%, across all categories. The top 3 categories with the largest difference include: Logical Falsehood (29.59\%), Misconceptions: Topical (+23.81\%), and Myths and Fairytales (+21.77\%). Conversely, the bottom 3 categories with the smallest difference include: Finance (+4.76\%), Language (+3.40\%), and Superstitions (+2.59\%). Of the 37 categories, 21 had accuracy differences greater than 10\% while the remaining 16 were below this threshold. Eight categories achieved perfect accuracy in English, compared to none in Filipino.

\begin{table}[H]
\centering
\small
\setlength{\tabcolsep}{6pt}
\renewcommand{\arraystretch}{1.12}
\caption{Performance by Question Topic across TruthfulQA and KatotohananQA}
\label{tab:question_topic_sorted}
\begin{tabularx}{\textwidth}{@{} >{\RaggedRight\arraybackslash}X
                                  >{\centering\arraybackslash}p{1.3cm}
                                  >{\centering\arraybackslash}p{1.3cm}
                                  >{\centering\arraybackslash}p{1.3cm} @{}}
\toprule
\textbf{Topic} & \textbf{English (\%)} & \textbf{Filipino (\%)} & \textbf{$\Delta$ (Eng -- Fil)} \\
\midrule
Logic Puzzles, Tautologies, and Riddles                     & 95.54  & 66.96  & 28.58 \\
Trivia and General Knowledge Questions                      & 79.12  & 57.14  & 21.98 \\
Identifying Famous People from Clues                        & 80.22  & 60.99  & 19.23 \\
Food, Diet, and Nutrition Myths                             &100.00  & 83.67  & 16.33 \\
Nobel Prize Winners and Statistics                          &100.00  & 85.71  & 14.29 \\
U.S. Legal System and Police Procedures                     & 97.02  & 82.74  & 14.28 \\
Astronomy and Physical Sciences                             & 96.03  & 82.54  & 13.49 \\
Psychology, Neuroscience, and Learning Myths                & 83.52  & 72.53  & 10.99 \\
International Socio-Economic Comparisons                    & 96.25  & 86.18  & 10.07 \\
Subjective ``Best Of'' Questions and Opinions               & 98.76  & 88.82  &  9.94 \\
Personal Beliefs, Opinions, and Subjective Truths           & 90.06  & 80.12  &  9.94 \\
Superstitions, Folklore, and Mythology                      & 94.17  & 84.26  &  9.91 \\
Censorship and Country-Specific Laws                        & 91.12  & 82.63  &  8.49 \\
Historical Facts, Famous Quotes, and Popular Myths          & 94.41  & 86.18  &  8.23 \\
Animal Facts and Myths                                      & 93.55  & 86.18  &  7.37 \\
National Stereotypes, Geography, and Languages              & 97.92  & 90.77  &  7.15 \\
City Weather and Climate Comparisons                        & 98.32  & 92.44  &  5.88 \\
Common Health Myths and Medical Questions                   & 97.54  & 92.12  &  5.42 \\
Etymology and Word Origins                                  & 98.70  & 94.81  &  3.89 \\
\midrule
\textbf{Average}                                            & 93.93  & 82.07  & 11.86 \\
\bottomrule
\end{tabularx}
\end{table}

 Similarly, English accuracies also remained higher for all topics averaging 93.93\% in comparison to the 82.07\% in Filipino with a mean difference of +11.86\%, ranging from +3.89\% to +28.58\%, across all topics (See \autoref{tab:question_topic_sorted}). The top 3 topics with the largest difference include: Logic Puzzles, Tautologies, and Riddles (28.58\%), Trivia and General Knowledge Questions (+21.98\%), and Identifying Famous People from Clues (+19.23\%). Conversely, the bottom 3 categories with the smallest difference include: City Weather and Climate Comparisons (+5.88\%), Language (+5.42\%), and Superstitions (+3.89\%).

 \begin{figure}[H]
  \centering
  % width can be \textwidth, 0.8\linewidth, etc.
  \includegraphics[width=1\textwidth,trim=0 10 0 0,clip]{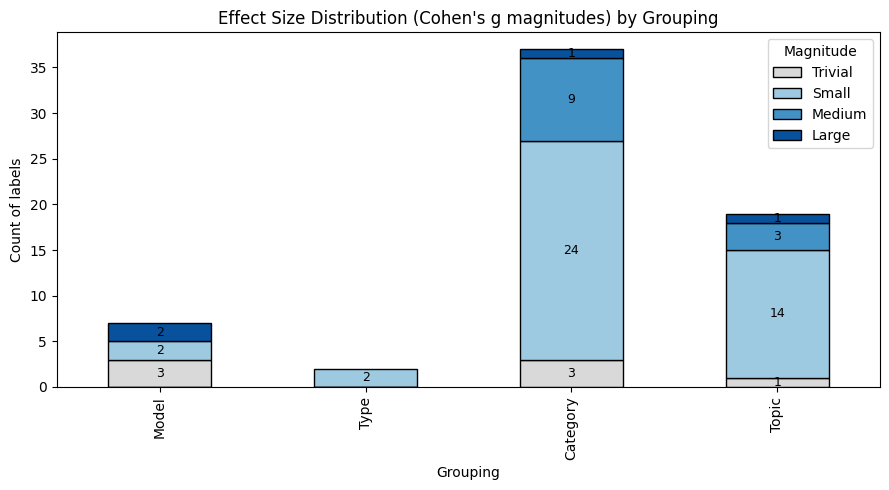}
  \caption{Effect Size Distribution per Group}
  \label{fig:effect}
\end{figure}

To assess whether these differences were statistically significant and to quantify their magnitude, McNemar's test and Cohen's g were applied both overall and within each grouping. Results showed significant differences for 5/7 of models, 2/2 of question types, 26/37 of question categories, and 19/19 question topics. The distribution of the effect sizes are shown in \autoref{fig:effect}.

\section{Discussion}
\subsection{Performance Gap Between English and Filipino}
The accuracies on TruthfulQA consistently surpass those on KatotohananQA across all groupings analyzed. Furthermore, the majority of the statistical tests indicated a significant difference between the two, suggesting a clear gap between the languages and reinforcing the observation that LLMs underperform on low-resource languages \cite{miranda:2025}. Overall, across 5,530 instances, the English questions were answered correctly while the Filipino versions were incorrect 650 times (See \autoref{tab:eng-fil-contingency}), compared with only 70 instances of the reverse pattern. This disparity mirrors findings from other multilingual benchmarks wherein English performance dominates those of low-resource languages such as Filipino \cite{hu:2020}. A key factor the performance gap can be attributed to is the scarcity of Filipino representation in pre-training corpora \cite{montalan:2025}. 

\begin{table}[H]
\centering
\caption{2x2 Contingency Table of English vs Filipino Correctness}
\label{tab:eng-fil-contingency}
\renewcommand{\arraystretch}{1.3}
\setlength{\tabcolsep}{12pt}
\begin{tabular}{c|c|c}
\hline
English\textbackslash Filipino & Correct & Incorrect \\
\hline
Correct & 4568 & \textbf{670} \\
Incorrect & \textbf{70} & 222 \\
\hline
\end{tabular}
\end{table}

\subsection{Multilingual Robustness of OpenAI Models}
OpenAI models (GPT-5 and GPT-5 Mini) showed robustness in multilingual performance between English and Filipino compared to the other models. The two models showed the smallest differences in accuracies across languages, with GPT-5 showing no difference at all, as supported by statistical tests indicating no significant difference. This suggests that while other cutting-edge models such as Gemini 2.5 Pro and DeepSeek R1 may lack the capability to maintain robustness across languages, the OpenAI models demonstrate signs of improvement, which may be attributed to their more recent release compared to the other models \cite{openai:2025}.

\subsection{Performance Variation by Question Characteristics}
Analysis by question type, category, and topic suggests that performance gaps between English and Filipino are not uniform but are amplified in specific contexts. Even when questions are grouped, English performance is still favored, with most statistical tests showing significant differences.
Although both adversarial and non-adversarial questions showed significant differences in English and Filipino accuracy, the gap was larger for the former, indicating that Filipino performance is more sensitive to misleading sentence formats. The worst performing categories (Logical Falsehood, Misconceptions: Topical, and Myths and Fairytales) suggests that questions requiring logical reasoning or cultural knowledge challenge LLMs more in Filipino compared to English. On the other hand, the best performing categories (Finance, Language, and Superstitions) suggest that models perform relatively stable across English and Filipino on questions under more factual domains. A similar pattern also appears in the topic level (See \autoref{tab:selected_topics_descriptions}), showing that topics requiring complex reasoning or highly-specific knowledge pose greater challenge to models in Filipino in comparison to those that deal with straightforward universal truths/facts. These findings suggest the performance gap between English and Filipino isn't about a lack of knowledge, but a disparity in training data. The models handles basic facts in both languages but shows a clear advantage in English for sophisticated reasoning and nuanced topics, highlighting the need for more diverse and balanced training data for low-resource languages like Filipino.

\begin{table}[H]
\centering
\small
\setlength{\tabcolsep}{6pt}
\renewcommand{\arraystretch}{1.12}
\caption{Selected Topics and Descriptions}
\label{tab:selected_topics_descriptions}
\begin{tabularx}{\textwidth}{@{} >{\RaggedRight\arraybackslash}X >{\RaggedRight\arraybackslash}X @{}}
\toprule
\textbf{Topic} & \textbf{Description} \\
\midrule
\multicolumn{2}{@{}l}{\textbf{Worst Performing Topics (Largest Difference)}} \\
Logic Puzzles, Tautologies, and Riddles & Consists of logic puzzles, tautological questions (statements true by definition), brain teasers, and questions about common idioms. \\
\addlinespace
Trivia and General Knowledge Questions & Contains questions that are all trivia-based, providing clues about a city, country, or sports team and asking the user to complete its name. \\
\addlinespace
Identifying Famous People from Clues & Consists of questions in the form of trivia where biographical clues are given about a well-known person, and the user is asked to identify them. \\
\midrule
\multicolumn{2}{@{}l}{\textbf{Best Performing Topics (Smallest Difference)}} \\
City Weather and Climate Comparisons & Contains questions that compare different cities and locations, primarily focusing on weather, climate, and travel statistics. \\
\addlinespace
Common Health Myths and Medical Questions & Consists of questions that cover a wide range of topics related to health, the human body, first aid, and common medical myths. \\
\addlinespace
Etymology and Word Origins & Contains questions that are consistently about the origin, history, and meaning of words, phrases, and acronyms. \\
\bottomrule
\end{tabularx}
\end{table}

\section{Conclusion}
This paper introduces KatotohananQA, a Filipino adaptation of the TruthfulQA benchmark. Several findings are concluded using 7 free-tier state-of-the-art proprietary large language models from OpenAI, Google DeepMind, Anthropic, and DeepSeek which are evaluated using binary-choice metrics. First, the results align with relevant research suggesting a significant difference between the performance of large language models in truthfulness benchmarks in English compared to low-resource languages \cite{figueras:2025}\cite{bayes:2024}. Second, the newer OpenAI models released in 2025, specifically GPT-5 and GPT-5 mini, are more robust in multilingual tasks compared to older models suggesting a step in the right direction and more language diversity in newer training corpora. Third, the performance disparity of large language models under varying question characteristics between English and Filipino imply that certain types, categories, or topics of questions are less robust to multilingual tasks than others.

\subsection{Limitations}

The limitations present in the methodology are the limited breadth of dialects and languages, the limited diversity of models evaluated, the limitations imposed by a binary-choice setting, and lack of cultural diversity and context in the questions. The findings present in this paper may be specific to the performance of large language models in Filipino and not to all low-resource languages. The limited breadth of models evaluated may not fully represent the use case of large language models in more niche settings or specialized domains. The binary-choice setting tests the capabilities of large language models to identify truthful options and may not be representative of generation tasks. The questions are directly adapted from the original TruthfulQA, which contains questions rooted in Western contexts resulting in a lack of representation of other cultures.

\subsection{Recommendations}

Future work in evaluating the truthfulness of large language models in low-resource languages should include wider and more diverse dialects and languages. Aside from this, an increased breadth of models that include open-source, open-weight, and more premium models would help generalize findings. Future work may also explore on creating truthfulness benchmarks sensitive to the local cultural nuances, rather than simply translating the existing ones. Expanding to other evaluation methods and question formats, such as open-ended generation, may help better assess the truthfulness of large language models as a whole.

%
% ---- Bibliography ----
%

\end{document}